\documentclass[sigconf]{aamas} 
\usepackage{balance} % for balancing columns on the final page

\setcopyright{ifaamas}
\acmConference[AAMAS '23]{Proc.\@ of the 22nd International Conference
on Autonomous Agents and Multiagent Systems (AAMAS 2023)}{May 29 -- June 2, 2023}
{London, United Kingdom}{A.~Ricci, W.~Yeoh, N.~Agmon, B.~An (eds.)}
\copyrightyear{2023}
\acmYear{2023}
\acmDOI{}
\acmPrice{}
\acmISBN{}

\acmSubmissionID{???}

\title[Stubborn: An Environment for Evaluating Stubbornness between Agents with Aligned Incentives]{Stubborn: An Environment for Evaluating Stubbornness between Agents with Aligned Incentives}

\author{Ram Rachum}
\additionalaffiliation{funding from ALTER, The Association For Long Term Existence And Resilience}
\affiliation{
  \institution{Bar-Ilan University}
  \city{Ramat Gan}
  \country{Israel}}
\email{rachumr@biu.ac.il}

\author{Yonatan Nakar}
\affiliation{
  \institution{Tel Aviv University}
  \city{Tel Aviv}
  \country{Israel}}
\email{yonatannakar@mail.tau.ac.il}

\author{Reuth Mirsky}
\affiliation{
  \institution{Bar-Ilan University}
  \city{Ramat Gan}
  \country{Israel}}
\email{mirskyr@cs.biu.ac.il}

\begin{abstract}

Recent research in multi-agent reinforcement learning (MARL) has shown success in learning social behavior and cooperation. Social dilemmas between agents in mixed-sum settings have been studied extensively, but there is little research into social dilemmas in fully-cooperative settings, where agents have no prospect of gaining reward at another agent's expense.

While fully-aligned interests are conducive to cooperation between agents, they do not guarantee it. We propose a measure of "stubbornness" between agents that aims to capture the human social behavior from which it takes its name: a disagreement that is gradually escalating and potentially disastrous. We would like to promote research into the tendency of agents to be stubborn, the reactions of counterpart agents, and the resulting social dynamics. 

In this paper we present \textit{Stubborn}\footnote{The code and installation instructions for Stubborn are available at \url{https://github.com/cool-RR/stubborn}}, an environment for evaluating stubbornness between agents with fully-aligned incentives. In our preliminary results, the agents learn to use their partner's stubbornness as a signal for improving the choices that they make in the environment.

\end{abstract}


\newcommand{\BibTeX}{\rm B\kern-.05em{\sc i\kern-.025em b}\kern-.08em\TeX}

\begin{document}

%%% The following commands remove the headers in your paper. For final 
%%% papers, these will be inserted during the pagination process.

\pagestyle{fancy}
\fancyhead{}

%%% The next command prints the information defined in the preamble.

\maketitle 

%%%%%%%%%%%%%%%%%%%%%%%%%%%%%%%%%%%%%%%%%%%%%%%%%%%%%%%%%%%%%%%%%%%%%%%%

\section{Introduction}

Mirsky and Stone \cite{MirskyStone2021} pose a challenge of using AI to produce an autonomous robot for assisting visually impaired people, similarly to a seeing-eye dog. Among the discussion of the technological, ethical and social challenges involved in this project, we would like to focus on the issue of \textit{intelligent disobedience}\cite{MirskyStone2021Asimov}.

When a flesh-and-bones seeing-eye dog is assisting its human operator in walking down a street, one of its tasks is to notice obstructions and dangers, and communicate with its operator accordingly. In the case of a full obstruction, or a passing vehicle, the seeing-eye dog is highly motivated to physically block the operator from proceeding, even when the operator is pleading and pushing the seeing-eye dog to move forward. The seeing-eye dog cannot communicate verbally, and the visually-impaired operator cannot see the obstruction or danger; therefore the only way for the operator to gauge the severity of the obstruction or danger is to escalate their pleading and pushing, and to observe how far the seeing-eye dog will go in matching their insistence. This tug of war, while frustrating for both human and dog, serves as a means of communication between the two.

In this scenario, the two agents are in conflict not because they are competing for an exclusive resource, but because they are both motivated to reach the best possible agreement for their mutual benefit. Unlike most sequential social dilemmas\cite{SSD} studied with multi-agent reinforcement learning, it results from a divergence of observations rather than motivations\cite{InequityAversion}. 

In this paper we present \textit{Stubborn}, an environment for evaluating stubbornness between agents with fully-aligned incentives. The rules for this environment are a simplified version of the dilemma described above. In our preliminary results, the agents learn to use their partner's stubbornness as a signal for improving their choices in the environment.

\section{Environment rules}

In the Stubborn environment, two agents are presented with two possible rewards, and they have to make a unanimous choice for one of these rewards. We explore the tension between the agents' motivation to choose the biggest reward, and their motivation to reach an agreement quickly.

\begin{enumerate}
    \item The environment is fully cooperative; the reward for Agent A is always equal to the reward for Agent B.
    \item The environment is partially observable. Crucially, the agents' observations are different from each other: Each agent observes an estimate of the left reward and an estimate of the right reward. Each estimate is sampled from a normal distribution around the true reward. The rewards themselves are sampled from a uniform distribution between 0 and 10.
    \item Each agent also observes the moves taken by the other player.
    \item Each episode of the game consists of $N_t$ turns with simultaneous actions. In each turn, each agent chooses whether to take the left reward or the right reward. If both agents choose the same reward, they both get the true value of the reward they've chosen. If they choose different rewards, they both get a reward of zero, which is always less than even the smallest reward, and continue playing until they reach an agreement or until the episode has reached $N_t$ turns.
    \item When the agents agree on a reward, two new random rewards are generated, along with four new estimates. The agents then play again for these rewards. Each sequence of turns is called a "skirmish", i.e. whenever the agents reach an agreement, a new skirmish is started.
    \item If both agents happen to change their choice at the same time, we pick one of the rewards at random and give it to both of them. In practice this means that each agent has a surefire way of ending any disagreement immediately.
\end{enumerate}

A skirmish can last anywhere between 1 and $N_t$ turns. A trivial example of a skirmish: Both agents immediately choose the left reward, and the skirmish ends with a length of one turn. Both agents get the left reward and a new skirmish begins. A non-trivial example of a skirmish: On the first turn Agent A chooses left while Agent B chooses right. On the second turn Agent A chooses left again while Agent B chooses right again. On the third turn, both agents choose right. Both agents get the right reward and a new skirmish begins.

Whenever agents disagree about which reward to take, they get a reward of zero. Intuitively, agents want to choose the biggest reward, but they would also like to prevent a long and costly disagreement with their partner. Crucially, when agents enter into conflict with each other, they don't know in advance how long the conflict will last, and which agent will give in first. The bigger the difference that an agent perceives between the two rewards, the bigger is its temptation to insist and hope that its partner will come around.

\section{Preliminary results}

We trained two agents with separate models on this environment, using a Proximal Policy Optimization (PPO) algorithm\cite{Schulman2017} provided by the RLlib framework\cite{RLlib}. We trained the two agents for 5,000 generations, with each episode consisting of 40 turns.

\begin{figure}[h]
    \includegraphics[width=\linewidth]{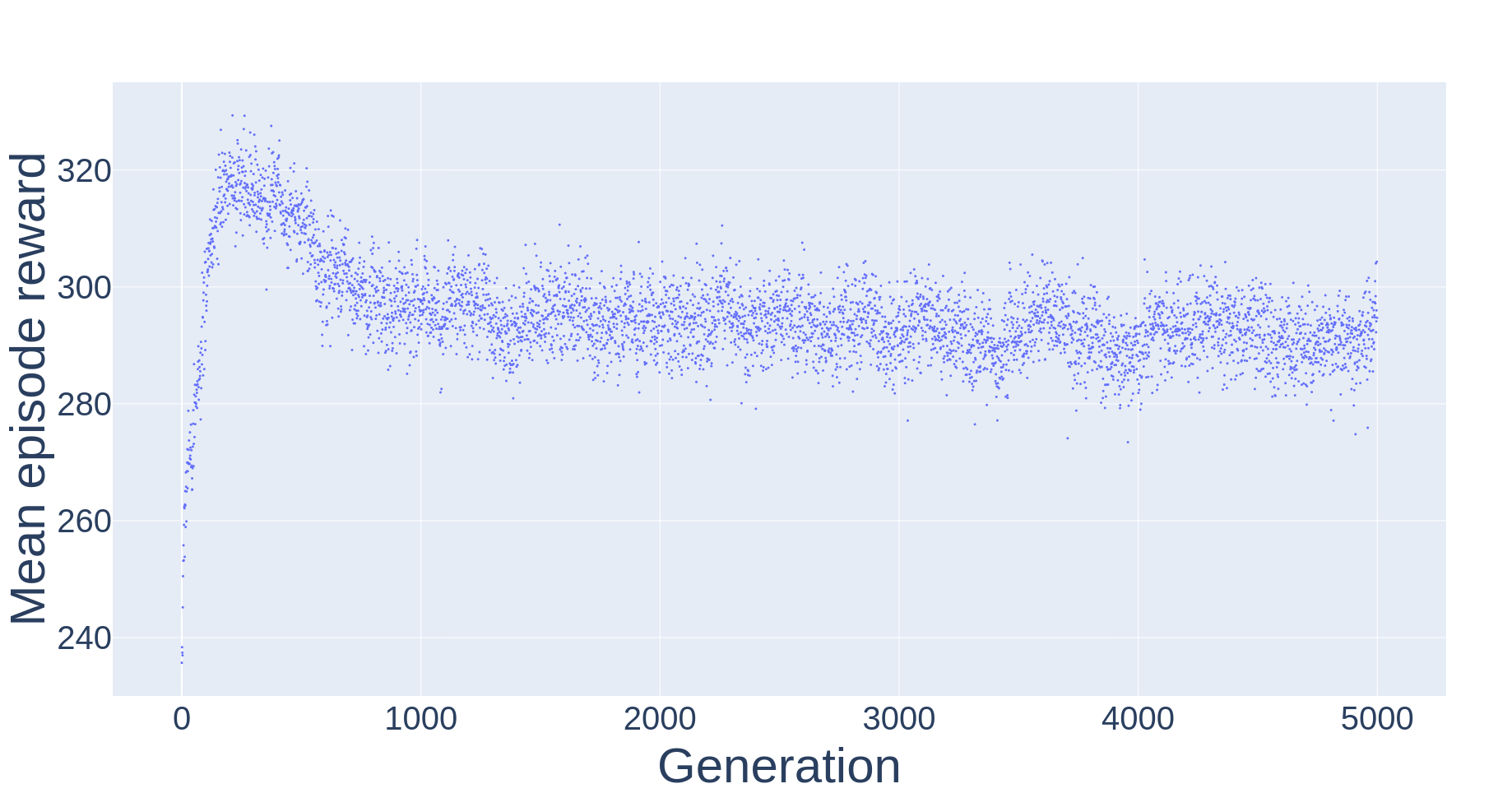}
    \centering
    \caption{Agent reward over 5,000 training generations}
    \label{fig:reward}
\end{figure}

We observe in Figure \ref{fig:reward} that the agents converge on a reward that's slightly higher than the initial reward.

We would like to measure how the stubbornness of one agent affects the decision-making of its counterpart. For each training generation, we perform a counterfactual measurement on the behavior of the trained policies. We define $\zeta_{n,d}$ as the probability that an agent will choose the left reward, where:

\begin{enumerate}
    \item According to its estimate, the left reward is $d$ points higher than the right reward.
    \item The two agents spent the last $n$ turns disagreeing, with the given agent choosing the left reward, and the other agent choosing the right reward.
\end{enumerate}

Intuitively, $\zeta_{n,d}$ represents the likelihood of the agent to insist on its initial choice, extending the skirmish by an unpredictable number of turns.

In Figure \ref{fig:zeta}, we show the average value of $\zeta_{n,d}$ over 5,000 training generations, with $n\in{0, 1, 2, 3, 4}$ and $d=5$:

\begin{figure}[h]
    \includegraphics[width=\linewidth]{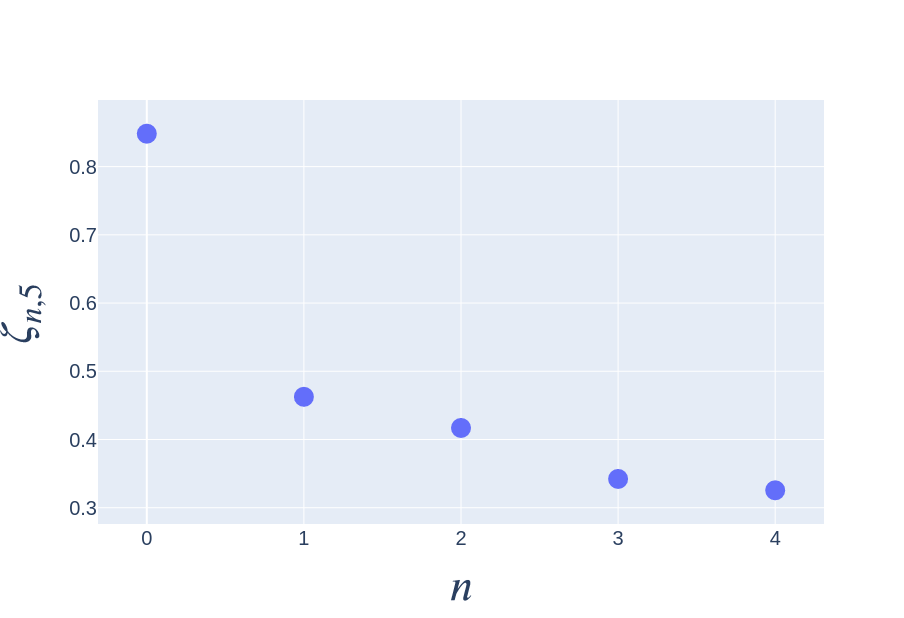}
    \centering
    \caption{Average value of $\zeta_{n,d}$ over 5,000 training generations, with $n\in{0, 1, 2, 3, 4}$ and $d=5$}
    \label{fig:zeta}
\end{figure}

We observe an inverse correlation: The more an agent insists on a certain choice, the more likely its counterpart agent is to agree to change its choice, even if that choice is seen as inferior according to its estimates. The length of the conflict, which is proportional to the sunk opportunity cost incurred by it, is used as a factor in the agent's decision whether to prolong the skirmish or end it.

\section{Discussion and future work}

We presented a new measure for fully-cooperative MARL, in which we quantify how stubborn the agents are. We show that in a single use case, each agent's stubbornness was correlated to the other agent's likelihood to yield to them.

We initially experimented with a more general variation of the environment rules listed above. Specifically, we set up the environment so one agent's estimates are more accurate than its partner. We defined a \textit{handicap} for each agent, which is a number that determines the variance of the normal distribution of the estimate. For example, an agent with a handicap of $2$ would get more accurate estimates than an agent with a handicap of $7$. Moreover, we experimented with setting the handicaps of both agents to a random number on each episode, while each agent knows its own handicap but not its partner. We hoped that the agents would use their stubbornness as a means of communicating the size of their handicaps to each other; if both agents know that Agent A has more accurate estimates than Agent B, a social convention where Agent B always yields would be beneficial for both agents. This behavior did not occur, but it's possible that a variation on this idea would converge.

Joining efforts with Boggs et al. \cite{Boggs2018} to provide tools for RaD-AI researchers to reach new insights, we've made our code available under the MIT open-source license. We hope to promote more research into this scenario.

\section{Acknowledgements}

We thank Edgar A. Duéñez-Guzmán and Georg Ostrovski for their valuable insights and suggestions that helped us refine the ideas presented in this work.

%%%%%%%%%%%%%%%%%%%%%%%%%%%%%%%%%%%%%%%%%%%%%%%%%%%%%%%%%%%%%%%%%%%%%%%%

%%% The next two lines define, first, the bibliography style to be 
%%% applied, and, second, the bibliography file to be used.

\bibliographystyle{ACM-Reference-Format} 
\bibliography{bibliography}

%%%%%%%%%%%%%%%%%%%%%%%%%%%%%%%%%%%%%%%%%%%%%%%%%%%%%%%%%%%%%%%%%%%%%%%%

\end{document}